%% file: novelty.tex
\relax
\documentclass[letterpaper]{article} 
\usepackage{arxiv} 
\usepackage{times}  
\usepackage{helvet} 
\usepackage{courier}  
\usepackage[hyphens]{url}  
\usepackage{graphicx} 

\usepackage{natbib}  
\usepackage{caption} 
 \usepackage{multirow}
\usepackage{times}
\usepackage{epsfig}
\usepackage{graphicx}
\usepackage{algpseudocode}
\usepackage{algorithm}
\usepackage{amsmath}
\usepackage{enumitem}
\usepackage{amssymb}
\usepackage{amsthm}
\usepackage{microtype}
\newtheoremstyle{mydefstyle}{}{}{\itshape}{}{\bfseries}{:}{.5em}{#1 #2 (\thmnote{#3})}
\theoremstyle{mydefstyle}

\def\real{\mathbb{R}}

\long\def\advercomment#1{}
\long\def\comment#1{}
\newcommand{\E}{\mathbb{E}}

\begin{document}

\title{A Unifying Framework for Formal Theories of Novelty:Framework,  Examples and Discussion }
\author {
    T. E. Boult\textsuperscript{\rm 1},
    P. A. Grabowicz\textsuperscript{\rm 5},
    D. S. Prijatelj\textsuperscript{\rm 2},
    R. Stern\textsuperscript{\rm 6},
    L. Holder\textsuperscript{\rm 4},
    J. Alspector\textsuperscript{\rm 3},\\
    M. Jafarzadeh\textsuperscript{\rm 1},
    T. Ahmad\textsuperscript{\rm 1},
    A. R. Dhamija\textsuperscript{\rm 1},
    A. Shrivastava\textsuperscript{\rm 7},
    C. Vondrick\textsuperscript{\rm 8},
    W. J. Scheirer\textsuperscript{\rm 2} \\
    \textsuperscript{\rm 1} U. Col. Col. Springs,
    \textsuperscript{\rm 2} U. Notre Dame,
    \textsuperscript{\rm 3} IDA/ITSD,
    \textsuperscript{\rm 4} Wash. State U.\\,
    \textsuperscript{\rm 5} U. Mass.,
    \textsuperscript{\rm 6} PARC, BGU,
    \textsuperscript{\rm 7} U. Maryland,
    \textsuperscript{\rm 8} Columbia U.\\
    \{tboult $\vert$ mjafarzadeh $\vert$ tahmad $\vert$ adhamija\}@vast.uccs.edu \\ grabowicz@cs.umass.edu  derek.prijatelj@nd.edu  rstern@parc.com  jalspect@ida.org  holder@wsu.edu\\  abhinav@cs.umd.edu vondrick@cs.columbia.edu  walter.scheirer@nd.edu
}

\maketitle


  

\begin{abstract}



Managing inputs that are novel, unknown, or out-of-distribution is critical as an agent moves from the lab to the open world. Novelty-related problems include being tolerant to novel perturbations of the normal input, detecting when the input includes novel items, and adapting to novel inputs.  While significant research has been undertaken in these areas, a noticeable gap exists in the lack of a formalized definition of novelty that transcends problem domains. As a team of researchers spanning multiple research groups and different domains,  we have seen, first hand, the difficulties that arise from ill-specified novelty problems, as well as inconsistent definitions and terminology. Therefore, we present the first unified framework for formal theories of novelty and use the framework to formally define a family of novelty types. Our framework can be applied across a wide range of domains, from symbolic AI to reinforcement learning, and beyond to open world image recognition. Thus, it can be used to help kick-start new research efforts and accelerate ongoing work on these important novelty-related problems.  

\end{abstract}

\section{Introduction} 

``What is novel?" is an important AI research question that informs the design of agents tolerant to novel inputs. 
Is a noticeable change in the world that does not impact an agent's task performance a novelty? 
How about a change that impacts performance but is not directly perceptible?  
If the world has not changed but the agent senses a random error that produces an input that leads to an unexpected state, is that novel?  

With decades of work and thousands of papers covering novelty detection and related research in anomaly detection, out-of-distribution detection, open set recognition, and open world recognition, one would think that a consistent unified definition of novelty would have been developed.  Unfortunately, that is not the case. Instead, we find a plethora of variations on this theme, as well as \textit{ad hoc} use and inconsistent reuse of terminology, all of which injects confusion as researchers discuss these topics.   

This paper introduces a unifying formal framework of novelty. The framework seeks to formalize what it means for an input to be a novelty in the context of agents in artificial intelligence or in other learning-based systems.
Using the proposed framework, we formally define multiple types of novelty an agent can encounter.
The goal of these definitions is to be broad enough to encompass and unify the full range of novelty models that have been proposed in the literature~\cite{pimentel2014review,markou2003novelty,markou2003novelty2,openset-pami13, openworld_2015,langley2020open}. An important generalization beyond prior work is that we consider novelty in the world, observed space, and agent space (see Fig.~\ref{fig:elements}), with dissimilarity and regret operators critical to our definitions.
The overarching goal is a framework such that researchers have clear definitions for the development of agents that must handle novelty, including support for agents / algorithms that incrementally learn from novel inputs.   A longer version of this theory with example applications to three different domains can be found at~\cite{Boult-eta-al-novelty20}.

Our framework supports \textit{implicit theories of novelty}, meaning the definitions use functions to implicitly specify if something is novel. The framework does not require a way to generate novelties, but rather it provides functions that can be used to evaluate if a given input is novel. This is similar to how any 2D shape can be implicitly defined by a function $f(x,y)=0$, whether or not there is a procedure for generating the shape. 
We contend any constructive or generative theory of novelty~\cite{langley2020open} must be incomplete because the construction or generation of defined worlds, states, and any enumerable set of transformations between them form, by definition, a closed world. We note, however, that a constructive model can be consistent with our definition, but we do not require a constructive model.

\input noveltytheory.tex

\section{Conclusion }

We see three primary contributions of this formalization of novelty that will spur further research.
First, formalization forces one to specify (or intentionally disregard) the required items in the theory.
This can lead to insights about the problem and fill in knowledge gaps.
For example, when applying the theory to the CartPole problem, numerous unanticipated issues were highlighted, new predictions made, and new experiments validated the new insights.

Second, formalization provides a common language to define and compare models of novelty across problems.
The precision of terms reduces confusion, while the flexibility allows it to be applied to a wide range of problems.

Third, the formalization allows one to make predictions about where or why experiments incorporating some form of novelty might run into difficulties.
For example, when the world-level and perceptual-level dissimilarity assessments disagree, we predict novelty problems will be more difficult.
One example of difficulty is world-disparity using variables not represented in perceptual space.
Another is when there are many possible world labels, but the input is only assigned one label that is used for assessing world-level dissimilarity.
In this case, the theory predicts a greater difficulty with such novelty, especially if the assigned label is associated with a physically smaller aspect of the observation.

Biological intelligence has a remarkable capacity to generalize novel inputs with ease, yet artificial agents continue to struggle with this behavior.
It is our hope that the adoption and use of the framework proposed here leads to the development of more effective solutions for novelty management and to make agents more robust to novel changes in their world.

By formalizing CartPole using our novelty framework, we gained insights into what are meaningful ``novelty'' problems for this task.
We showed how to develop better measures to predict when novelty would be easy or hard to manage or to detect.
In line with this, our team of researchers has been refining this theory and applying it to multiple problem domains.
More details can be found in the longer arXiv version~\cite{Boult-eta-al-novelty20}.

{\footnotesize
\paragraph*{Acknowledgment} 
This research was  sponsored  by the Defense Advanced Research Projects Agency (DARPA) and the Army Research Office (ARO) under multiple contracts/agreements including  HR001120C0055, W911NF-20-2-0005,W911NF-20-2-0004,HQ0034-19-D-0001, W911NF2020009. The views contained in this document are those of the authors and should not be interpreted as representing the official policies, either expressed or implied, of the DARPA or ARO, or the U.S. Government.
}

{\small
\bibliography{novelty}
}

\end{document}

%% file: noveltytheory.tex
\section{Formalizing Novelty} 

\abovedisplayskip=4pt plus 2pt minus 1pt 
\abovedisplayshortskip=2pt plus 1pt minus 1pt 
\belowdisplayskip=4pt plus 2pt minus 1pt 
\belowdisplayshortskip=2pt plus 1pt minus 1pt

\begin{figure}[t]
 \centering
    \includegraphics[width=\columnwidth]{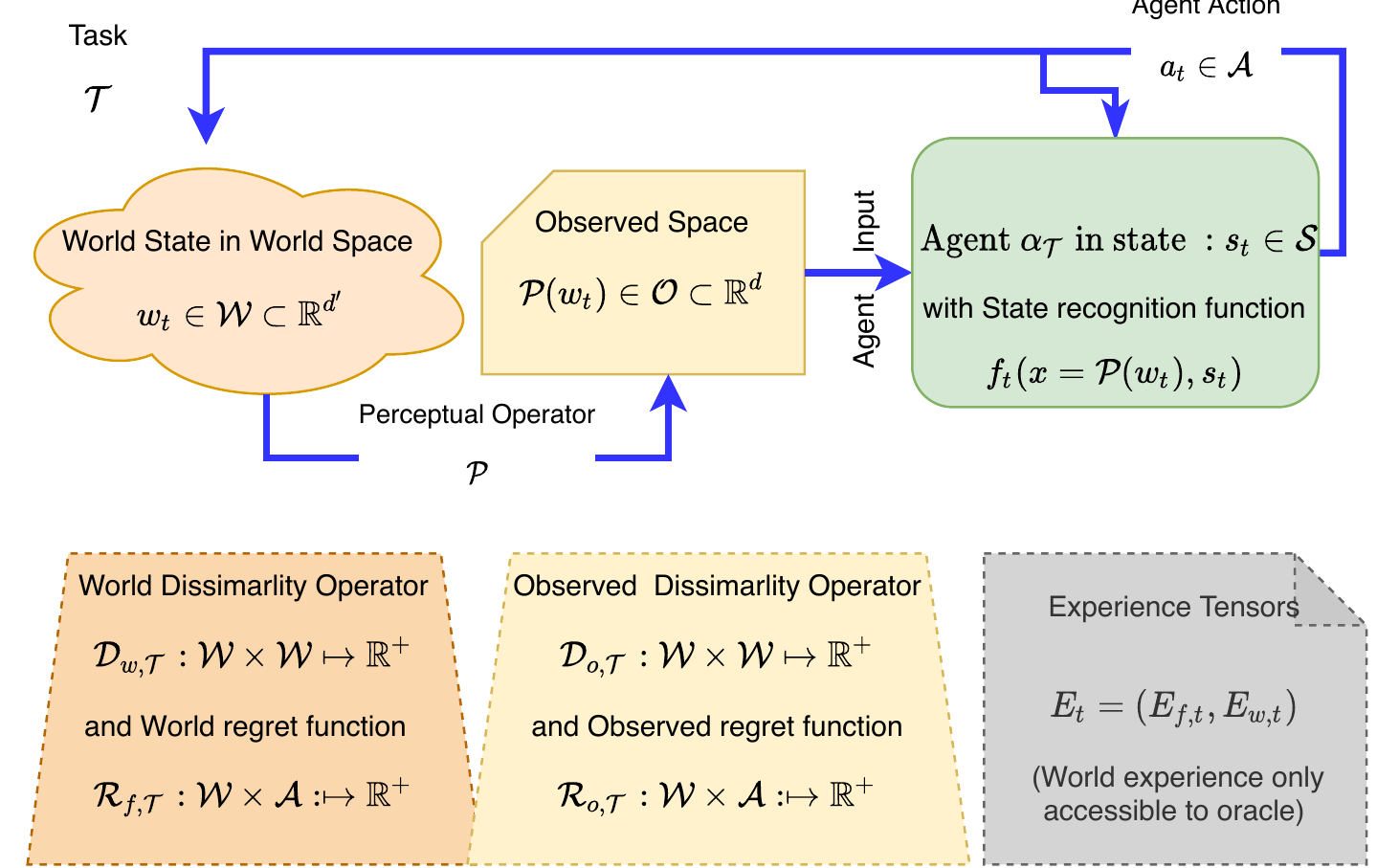}
   \caption{Main elements of the implicit theories of novelty. The agent can only access world information indirectly through a perceptual operator ${\cal P}$. It can then update its internal state and act on the world state. Items with dashed outlines are outside of the task or agent but are critical to defining novelty.
In the framework a theory of novelty is obtained by specifying:
world $\cal W$ and the world dimensionality $d'$,
observation space ${\cal O}$ accessible to the agent and its dimensionality $d$, and agent state space $\cal S$
family of perceptual operators ${\cal P}$ with associated ${\cal W}_t$ processed world regions,
task-dependent world dissimilarity functions ${\cal D}_{w,{\cal T}; E_t}$ with associated threshold $\delta_w$,
task-dependent observation-space dissimilarity functions ${\cal D}_{o,{\cal T}; E_t}$ with threshold $\delta_o$,
agent $\alpha$ in state $s_t\in{\cal S}$ at time $t$, using 
state recognition function $f_t(x,s)$ to determine the action $a_t \in{\cal A} $ to be taken, 
world regret function ${\cal R}_{w,{\cal T}}$, 
observation-space regret function ${\cal R}_{o,{\cal T}}$.
and agent-space regret function ${\cal R}_{a,{\cal T}}$. 
Every  set of these operators / functions / values defines a different theory of novelty for its associated task. 
}
\label{fig:elements}
\end{figure}

We present frameworks for formalizing novelty for static or learning-based agents, operating in a setting where handling unknown items is required.  Fig.~\ref{fig:elements} shows the main elements of a novelty problem for task $\cal T$. 
 The formulation can support a wide range of novelty problems including being robust to novelties, detecting novelties, learning from novelties or generating novelties.   
 The paper's formalization is about theories, rather than ``a theory,'' because when the definition's set of items and associated functions are provided, a different theory of novelty is defined.  There are infinitely many such theories of novelty for any given task.

For simplicity of presentation, our world and observation space are $d'$ and $d$ dimensional spaces of real numbers. Let a world state be $w_t\in {\cal W}$, and let ${\cal W} \subset \real^{d'}, d'\ge d$, be the representation of the world state at time $t$ obtained from a subset of ${\cal W}$, the allowed world states. 
Note that $\cal W$ need not be the entire world. It can be only that region of space and time which is sampled during operation or that is relevant for the task. It could even be a finite set of items or relationships to be considered by the system.

Let our observation at time $t$ be $x_t\in {\cal O}$ in observation-space ${\cal O} \subset \real^d $. 
Let ${\cal P}:\real^{d'} \mapsto \real^d$ be the perceptual operator at time $t$, which maps world spaces to observation spaces, \textit{i.e.}, ${x_t = {\cal P}}(w_t)$.  The agent never has direct access to the world and can only access it via the perceptual operator. 
This operator is generally a combination of real-time sensing plus external processing on that sensed data. It can also include pre-processing on stored sensory data. But everything in this operator is to be considered external to the agent. Since system hardware can change over time, this mapping is a function of time. If the perceptual operator processes only a subregion of any world state, we let ${\cal W}_t$ be the subregion of the world that has been processed up to time $t$. Accordingly, any world states that differ only outside the processed subregions are indistinguishable.

Let agent $\alpha_{{\cal T}}$ solving task ${\cal T}$ at time $t$ have an internal state representation $s_t \in {\cal S}$, where ${\cal S}$ is the space of possible states.
An agent reacts to the environment, but a common architecture in the design of autonomous agents is for the agent to have an internal state $s_t$ that influences how the agent will act at time $t$. 
To capture this common agent architecture, we define a \emph{state recognition function} that maps an observation-space input $x_t$ plus the current agent state into its new internal state. 
Formally, let $f_t(\cdot,\cdot ):\real^d\times{\cal S}\ \mapsto {\cal S}\times {\cal A}$ be a state recognition function at time $t$ mapping an observation-space input $x_t$ to its internal state space estimation $y$ of the world state $w_t$ yielding the  action $a_t \in {\cal A}$ to be taken to reach new state $s_t$, where $\cal A$ is the space of actions.

Based on the recognition function output, the agent takes an action. Then, the state transition function,
$T(w_t,a_t):{\cal W} \times {\cal A} \mapsto \cal W$, maps the current world state and agent's selected action at time $t$ to a new world state. 
This mapping can be stochastic, e.g., a Markov decision process.
Note for many real problems the world may change outside of any assumptions, including oracle's assumptions, in which case we cannot assume $w_{t+1} = T(w_t,a_t)$.
For problems which traditionally do not have state
or specify an action, such as machine learning or open set classification, the agent ``action" is reporting the results
of the recognition function. Actions in learning systems may also update the parameters in the state or may
change state recognition functions, which is why $f_t$ is a function of time~$t$.

Let $\cal N\subseteq \cal S$ be the possible empty set of internal states which are associated with the agent determining the world is novel. The world state $w_t$ is identified by the agent as novel if 
$s_t \in {\cal N}$.  When dealing with novelty, the agent may obtain unexpected observation states, which could map to potentially unexpected internal states (\textit{e.g.}, a ``crash'' state).

To represent history, let $E_{t} = \{(w_1) \ldots (w_t)\}$ be our experience tensor of states. 
It is important to note that the experience tensor is not about what the agent ``remembers.‘’
 It is external to the agent and is about what world states the agent has experienced up to and including time $t$. 
The experience tensor is integral to defining novelty, which depends on dissimilarity between the world and experience.


\section{Dissimilarity and Regret Measures}
For a given task ${\cal T}$, data generally do not need to match exactly to be considered ``the same'' with respect to the task's objectives.  Note that while we call it ``a task,'' it could include a set of objectives (\textit{e.g.}, a multi-task problem). In any task, multiple dimensions can impact the task performance and determine when a state is effectively the same or how different a state is from prior experience.  We formulate this in terms of a measure of dissimilarity, which depends strongly on the task. It is important to note that one should be careful in {\em a priori} definitions of what matters to a task, as novel world states may have an unpredictable impact on it. 

Essential to our framework is a set of task-dependent dissimilarity functions ${\cal D}_{w,{\cal T};E_t}:{\cal W}\times {\cal W} \mapsto \real^+$ and ${\cal D}_{o,{\cal T};E_t}:{\cal W}\times {\cal W} \mapsto \real^+$, which measure dissimilarity between items in the world space and the observation space respectively.  The perceptual dissimilarity may access the world but must map to observed space with the perceptual operator before mapping to $\real^+$. 
Dissimilarity measures produce non-negative values that are a generalization of distance metrics. It is possible to have zero dissimilarity for items that are identical in terms of the task, even if they are distinct items in the world / observation space.  Novelty definitions will generally include a similarity threshold of $\delta$ on the maximum allowed dissimilarity for items treated the same by the task, which depends on the task / user requirements for matching.
Dissimilarity may be an actual distance metric, a statistical measure, an information-theoretic measure, or such measures combined with ontological or hierarchical information.  If the world modeling is probabilistic, then the dissimilarity function may, but need not, be based on probabilistic computations.   
We use dissimilarity rather than distance since it is well known that human perception / recognition is non-metric~\cite{tversky1977features,scheirer2014good}.

Not all novel items are of interest or present a risk to the agent.
While we cannot know the risk of an unknown item until it becomes known, we can, after the fact, assign a regret score associated with new world / observed / agent state after the action $a_t(w_t,s_t)$. 
We let ${\cal R}_{f,{\cal T}}:({\cal O} \times {\cal A} )\mapsto\real$, ${\cal R}_{o,{\cal T}}:({\cal O}\times {\cal A} )\mapsto\real$ and ${\cal R}_{w,{\cal T}}:({\cal W} \times {\cal A} )\mapsto\real$ be the regret operators for task ${\cal T}$ in agent, observation, and world space respectively. Agent and observation regret must process the world state via the perceptual operator.  A suboptimal agent may have regret higher than observational regret, which should be defined with respect to an oracle or the best agent on observed data.   

We separate these three because it allows one to reason about regret in terms of specific models. Only an oracle that has access to ground-truth data has the ability to actually compute regret in world space. However, an agent can approximate regret in observation space, especially given the ground-truth answers. It is important to note that agent-computed regret can only be an approximation, even given the ground-truth action / outcome, because the optimal decision with limited data can still lead to bad outcomes. Hence, an agent might estimate regret even if there should be none.

\section{Example: CartPole Domain}

As an example of using the constructs defined above, consider  the 
CartPole in the OpenAI Gym.\footnote{\url{https://github.com/openai/gym/blob/master/gym/envs/classic_control/cartpole.py}} 
In this domain, a cart has a pole connected to it, and the task ${\cal T}$ is to push the cart left or right so as to prevent the attached pole from falling. 
The world state $w$ in the CartPole domain comprises the following real values (with default / initial values in parentheses): 
gravity $G~(9.8)$, 
mass of cart $M_c~(1.0)$, 
mass of pole per unit length $M_p~(0.1)$,
length of pole $L~(1.0)$,
force of push $F_p~(10.0)$,
horizontal force acting on the cart $F_h~(0)$,
min / max cart position $z^{\text{min}} (-2.4), z^{\text{max}} (+2.4)$,
min / max pole angle $\phi^{\text{min}}$ ($-12^{\circ}$), $\phi^{\text{max}}$ ($+12^{\circ} $),
time between state updates $\tau$ ($0.02$ seconds), 
start time $t$ (0). The
initial cart position $z_0$, 
 cart velocity $\dot{x}_0$, 
 pole angle $\phi_0$, 
and  pole angular velocity $\dot{\phi}_0$ are all i.i.d. random samples from  $[-0.05...0.05]$.
The perceptual operator ${\cal P}$ in this domain is a projection of the world state that returns only the cart position $z$, cart velocity $\dot{z}$, pole angle $\phi$, and pole angular velocity $\dot{\phi}$ as the 4D observed state vector  $x=(z, \dot{z}, \phi, \dot{\phi})$. 

Based on these world state features, the task is more precisely defined as: given $x$, select an action from the space of ${\cal A}=\{\text{Left},\text{Right}\}$  to maintain the cart position within the min / max cart position and maintain the pole angle within the min / max pole angle. Note that the last four features (cart position, cart velocity, pole angle, pole angular velocity) are determined via a  deterministic physics model based on the full world state combined with agent actions.

\subsection{Dissimilarity and Regret in CartPole}

State transitions in the CartPole domain are determined by the equations of motion and can be simulated in discrete time with numerical integration.
In this example we  assume transitions between observed states are Markovian, which simplifies presentation;   other theories for this domain could consider dissimilarity and regret in more general settings. 

The dissimilarity measure for CartPole might take a simple form (\textit{e.g.}, the Euclidean distance in the world or observed space).
However, Euclidean distance between world states is affected by factors  other than novelties,
including  the choice of units. It is also insensitive to the  variation in the impact of different variables on the state evolution or task outcome.  
Proper conditioning would reduce dependency on units and account for states that correspond to different samples from the same world (\textit{e.g.}, the same CartPole world with a different initial position of the pole is not considered novel).

To avoid these issues, we compare two worlds, $w$ and $\check{w}$, the states that proceed from a common observed state and action.
We consider an action of an optimal agent,~$a^*$, in the first world, $w$, and choose as the common observed states the states that the agent encounters,~$\check{x}_t$, in the second world, $\check{w}$.
Then, we average over all these states, including the initial observed state:
\begin{align*}
 {{\cal D}}_{o,{\cal T}}(w, \check{w}) =
(
 &{\cal P}(T(M(w,\check{x}_t),a^*_{t})) - \\
 &{\cal P}(T(M(\check{w},\check{x}_t),a^*_{t}) 
 )^2,
\end{align*}
where $M(w,x):{\cal W} \mapsto {\cal W}$ is a function that returns a  modified $w$ whose observed components are replaced with the values from $x$, such that ${\cal P}(M(w,x))=x$, while all other components remain unchanged. 
Overall, the dissimilarity measures the average distance between observed states in two different worlds that proceed from a common observed state and action,~$\check{x}_t$ and $a^*$.
The agent is optimal in the first world, while the trajectory is from the second world, so this dissimilarity measure can be seen as an expected state prediction error of the optimal agent trained in the first world and tested in the second world.
%
The world dissimilarity is defined analogously, except it does not use the perceptual operator. 

Note this is an asymmetric dissimilarity measure as the selected agent is optimal for the first world and need not be optimal for the second, and then marginalizes over the initial conditions and time in the second world.
Due to the conditioning, any pair of states from the same world will have zero dissimilarity. Furthermore, since these depend on the choice of ``optimal action'' $a^*_{t}$ from the first world,  it implicitly normalizes for how different dimensions (variables) impact the evolution of the world state. 
One can consider actions of an optimal reference agent under given conditions (\textit{e.g.}, a non-adaptive agent that performs optimally w.r.t. the main task $\cal T$ in a non-novel CartPole world.
If it is infeasible to obtain an optimal agent in practice, then one may use an arbitrary reference agent and its expectation, with the caveat the dissimilarity measure will depend on the oracle's reference agent.

Regret for the agent's action at state $x_t$ w.r.t. $\cal T$ is
\begin{align}
    {\cal R}_{o,{\cal T}} ( x_t, a_t) = 
    {\cal R}_{w,{\cal T}} ( w_t, a_t) = 
    \ell_{\cal T} (w_t, a_t) - \ell_{\cal T} (w_t, a^*_t),
     \nonumber
    \label{eq:Rx}
\end{align}
where $\ell_{\cal T} (w_t, a_t)$ is the loss incurred in the world $w_t$ by a given agent that observes state $x_t$ and performs action $a_t$, while $\ell_{\cal T} (w_t, a^*_t)$ is the loss of an agent that performs the optimal action $a^*_t$ in the world.
In CartPole, the loss at the given time step is $1$ if the pole angle or cart's position in the next time steps is beyond the threshold, $|\phi|>\phi^{\text{max}}$ or $|z|>z^\text{max}$, otherwise the loss is $0$.
In this CartPole domain, the world regret is the same as observation regret, because there are no hidden dynamic elements interacting with pole or cart, such as an invisible pendulum that hangs above the pole and sometimes hits it. Once such elements are introduced, the two regrets may differ.

\begin{figure}
    \centering
    \includegraphics[width=.49\columnwidth]{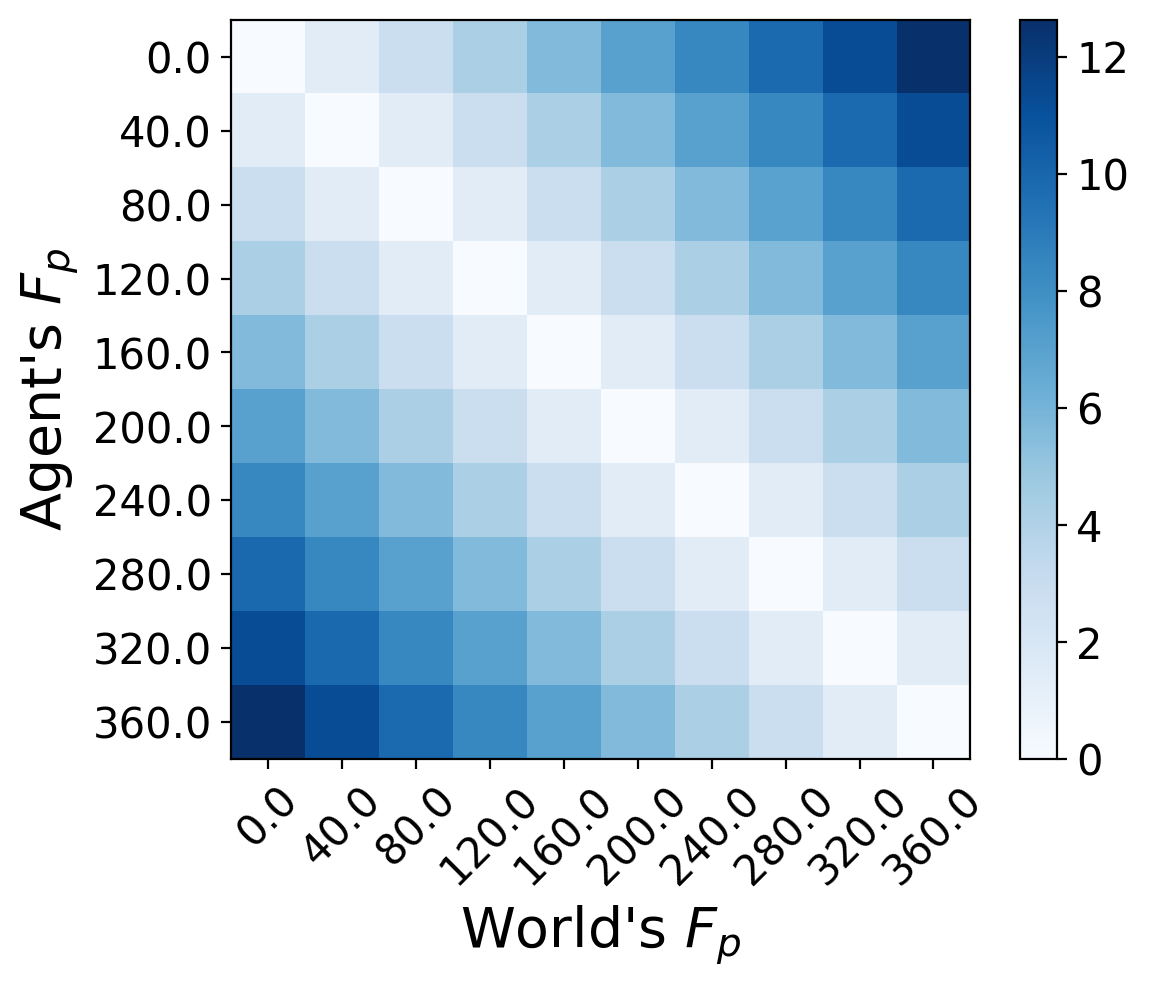}
    \includegraphics[width=.49\columnwidth]{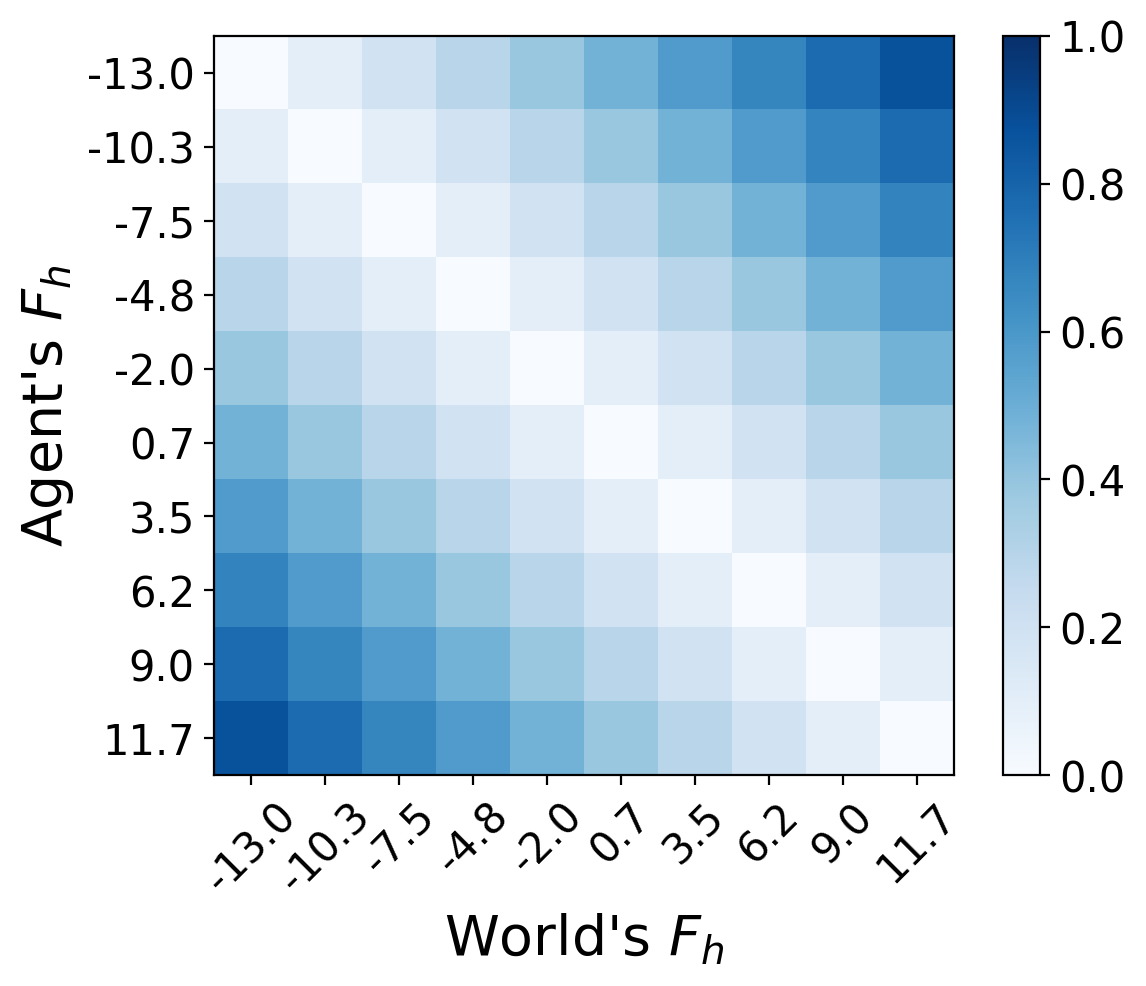}
    \caption{Average dissimilarity, $\E_{x_0',t} {{\cal D}}_{o,{\cal T}}(\check{w}_t, w_t)$, between the future states expected and observed by agents that were trained in a world with incorrect value of the magnitude of pushing force, $F_p$ (left panel), or a horizontal force acting on the cart, $F_h$ (right panel). The expectation is computed over 20 samples of the initial world state $w_0$.}
    \label{fig:diss}
\end{figure}

\begin{figure}
    \centering
    \includegraphics[width=.49\columnwidth]{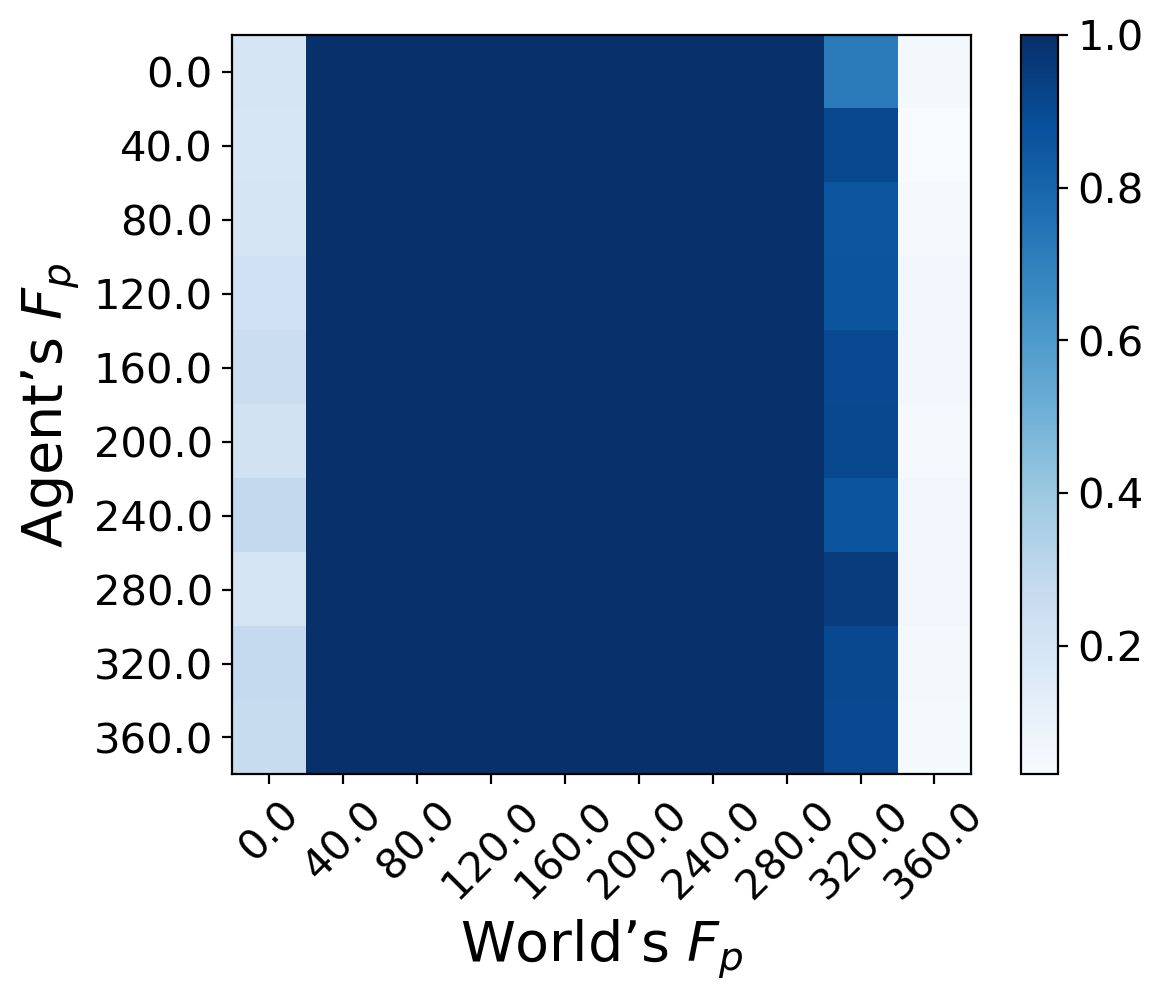}
    \includegraphics[width=.49\columnwidth]{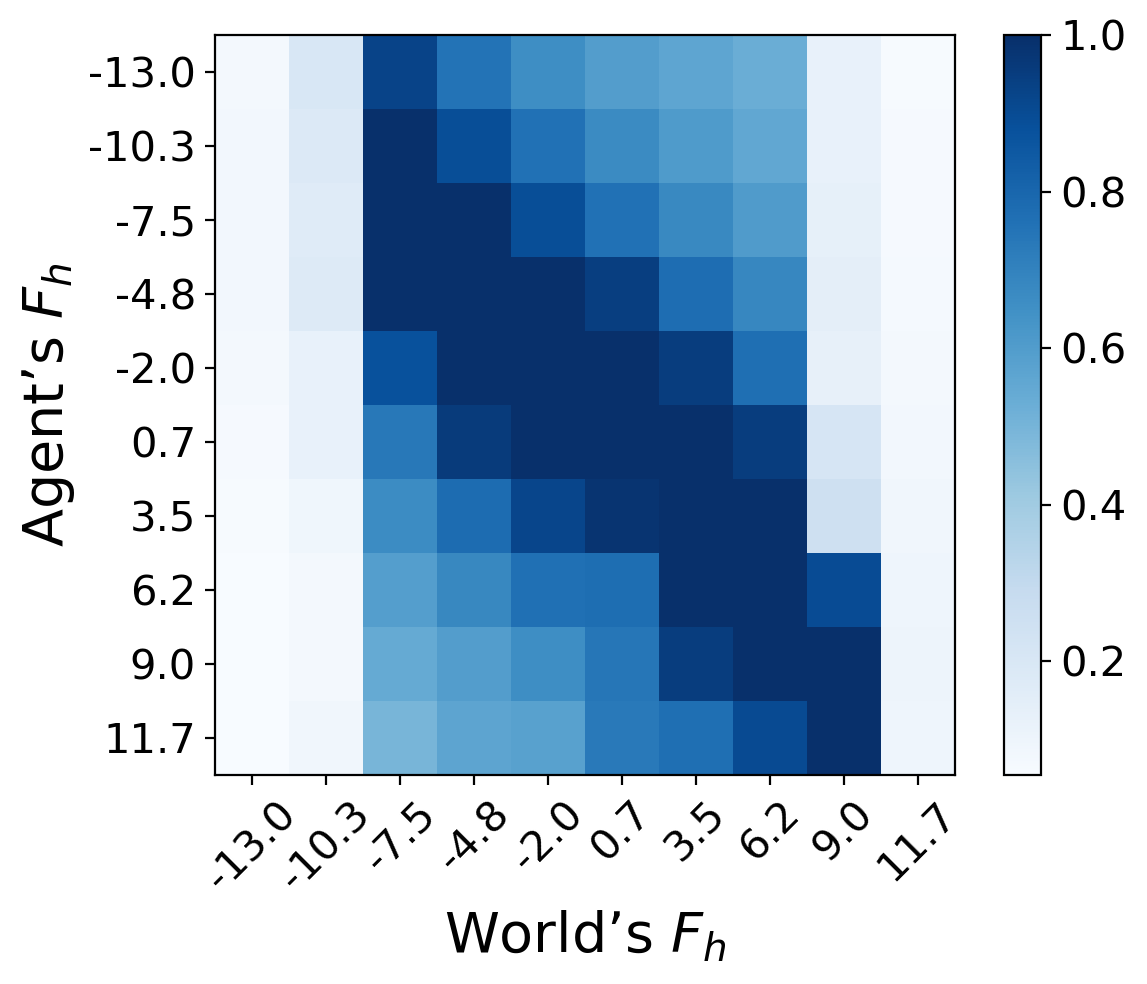}
    \caption{Average reward, 1-$\E_{w_0,t} \ell_{{\cal T}} ( w_t, a^*_t) $, of non-adaptive agents that are trained to act optimally in a certain world but are tested in another world.}
    \label{fig:regret}
\end{figure}

\subsection{Agent State and Optimal Non-Adaptive Agents}


In simple environments, like CartPole, it is possible to obtain optimal or near-optimal non-adaptive agents. A particularly simple version of CartPole is the one where the agent’s action space is binary, \textit{i.e.}, the agent can choose to push the cart left or right, and its reward is the time that the pole is up. In such a CartPole environment, an agent can find optimal actions by performing what-if simulations of the world and searching for actions that result in the best performance, \textit{i.e.}, the agent can simulate what would happen if it pushed the cart left or right and then choose the next action that results in better performance.  In this paper, for simplicity, we present the results for a near-optimal single-look-ahead agent. 
The action is chosen based on the distance, $||\beta^\intercal (x_t-x^\text{s})||$, between the expected state resulting from the action, $x_t$, and the desired state, $x^\text{s}=(0,0,0,0)$, where the weight vector $\beta=(0,0,1,0.005)$ weighs discrepancy in $\phi$ the most and ignores the discrepancies in $z$ and $\dot{z}$.
The state of this agent is described by the parameters used to simulate system dynamics and the weight vector, $s=(G, M_c, M_p, L, F_p, F_h, \tau, \beta)$. In non-adaptive agents, these parameters are fixed to the values that either correspond to the environment (no novelty) or do not (a novelty). 

\subsection{Measurements and Observations}

As expected, the dissimilarities in observed state prediction capture the novelties in the magnitude of pushing force and in a horizontal force acting on the cart (Figure~\ref{fig:diss}). The larger the distance between the value of a given parameter in the environment and its value assumed by the agent, the larger the dissimilarity in state prediction.

Surprisingly, in CartPole, reasonable changes to many of the aforementioned latent parameters, like the gravity or the magnitude of pushing force, do not impact an optimal non-adaptive agent’s performance (left Figure~\ref{fig:regret}). 
The columns on the verge of the heat map correspond to the achievable limits in performance, \textit{i.e.}, the leftmost column corresponds to such a small force magnitude that the push is insufficient to counter the gravity, whereas the rightmost column corresponds to such a large force magnitude that the push instantly rotates the pole beyond the allowed region.
We conclude that the optimal agent performs just as well in the novel world where gravity or pole length have new values, despite making simulations that assume incorrect values of gravity or pole length, \textit{i.e.}, the values from the non-novel world. 
Note that this agent is non-adaptive, so it does not change its internal model to adapt to the new environment, \textit{i.e.}, it has exactly the same internal model as in the non-novel environment.
Again, the columns at the verge of the heat map mark inherent performance limits, \textit{i.e.}, if the horizontal force is larger than the magnitude of the pushing force, $F_p=10$, then the pushes are insufficient to counteract the horizontal force and the pole is destined to fall, despite taking optimal actions.

Less surprisingly, some of the latent parameters impact the agent’s performance in an intuitive way (\textit{e.g.}, a horizontal force applied to the cart) (right Figure~\ref{fig:regret}). This happens because the horizontal force directly impacts the action that the agent should choose: if the horizontal force pushes the cart right, then the agent should probably push it left, and vice versa. The larger the difference between the horizontal force in the environment and assumed by the agent, the higher the drop in the agent's performance.

To distinguish between novelties that affect or do not affect the task performance, we use the regret of an optimal agent trained in the non-novel world and tested in the novel world, \textit{i.e.}, ${\cal R}_{o,{\cal T}} ( x_t, a^*_t) $. Novelties with ${\cal R}_{o,{\cal T}} ( x_t, a^*_t) = 0$ do not affect task performance and can be ignored by agents that are near-optimal without a performance drop.
By contrast, ${\cal R}_{o,{\cal T}} ( x_t, a^*_t) > 0$ tells us that the novelty impacts the performance of an optimal non-adaptive agent and that the agent can update its state to improve its performance, \textit{i.e.}, learn the novelty.
Naturally, one can develop an adaptive version of this agent by learning an estimate of world state parameters
from observations and using them to perform more accurate simulations and taking better actions. 

\if{0}
\subsection{Agent State}
The set of internal states ${\cal A}$ depends on the implementation of the agent. For our example, consider a simple DQN-based agent that runs for $N$ iterations to train a DNN to output the Q value of each action, 
and from iteration $N+1$ and onward chooses actions greedily according to learned network. 
Let DNN$_i$ be the trained network as iteration $i$. 
The internal state of the agent in iteration $i\leq N$ comprises all the observation-space states collected ($E_{o,t}$) and the so far and the current DNN (DNN$_i$). 
After the $N$th iteration, the agent's internal state is only the learned DNN, as the agent chooses its actions according to the DNN and the given observation-space state.   Formally, 
\begin{equation}
f_t(x : E_t) = 
\begin{cases}
(E_t, DNN_t) & \text{if $t\leq N$} \\
DNN_{100} & \text{if $t> N$} 
  \end{cases}
\end{equation}
The actions the agent can take at any time $t$ are ``left'' or ``right,'' which correspond to an instantaneous push of the cart in the corresponding direction with force $F_p$.
The world regret function $R_{w,T}$ is based on whether $x$ or $\phi$ exceed their bounds in $w$. This can be a Boolean (success / failure) or a score based on the amount of time that $x$ and $\phi$ remain within bounds. The observation space regret function $R_{f,T}$ has the same value as $R_{w,T}$, but is not necessarily determined by $f$. 
\fi

{
\renewcommand{\arraystretch}{1.15}
\begin{table*}[] \centering 
\begin{tabular}{|c|c|c|c|c|c|c|}
\multicolumn{1}{|c|}{\multirow{4}{*}{\rotatebox[origin=c]{90}{ \parbox{.2in}{World\\ Novelty}}}}&\multicolumn{1}{l|}{\multirow{2}{*}{\rotatebox[origin=c]{90}{\hspace{-6ex}\parbox{.2in} {Observation\\ Novelty}}}}& \multirow{2}{*}{\rotatebox[origin=c]{90}{\hspace{-6ex}\parbox{.2in}{Agent\\ Novelty}}} & 
\multicolumn{2}{c}{ World Regret ${\cal R}_{w,{\cal T}} > \epsilon_w$} & \multicolumn{2}{|c|}{{{No World Regret ${\cal R}_{w,{\cal T}} \le \epsilon_w$}}}          \\ \cline{4-7}
&&&\parbox{1.2in}{\vspace{2pt} \centering Perceptual Regret\\${\cal R}_{o,{\cal T}} > \epsilon_o$}
&\parbox{1.2in}{\vspace{2pt} \centering No Perceptual Regret\\${\cal R}_{o,{\cal T}} \le \epsilon_o$}
&\parbox{1.2in}{\vspace{2pt} \centering Perceptual Regret\\${\cal R}_{o,{\cal T}} > \epsilon_o$}      
&\parbox{1.2in}{\vspace{2pt} \centering No Perceptual Regret\\${\cal R}_{o,{\cal T}} \le \epsilon_o$}\\ \hline
\multicolumn{1}{|c|}{\multirow{4}{*}{\rotatebox[origin=c]{90}{{Yes }}}}     & \multicolumn{1}{l|}{\multirow{2}{*}{\rotatebox[origin=c]{90}{Yes}}}   & \multicolumn{1}{l|}{  Yes}  & \multicolumn{1}{l|}{Unanimous w/ Regret}                & \multicolumn{1}{l|}{Unanimous  Nuisance}      & \multicolumn{1}{l|}{Unanimous  Nuisance}      & \multicolumn{1}{l|}{Unanimous Managed}             \\ \cline{3-7} 
\multicolumn{1}{|c|}{}                                   & \multicolumn{1}{l|}{}                                      & \multicolumn{1}{l|}{  No} & \multicolumn{1}{l|}{Ignored}                  & \multicolumn{1}{l|}{Ignored Nuisance}         & \multicolumn{1}{l|}{Ignored Nuisance}         & \multicolumn{1}{l|}{Ignored Managed}               \\ \cline{2-7} 
\multicolumn{1}{|c|}{}                                   &
\multicolumn{1}{l|}{\multirow{2}{*}{\rotatebox[origin=c]{90}{No}}} 
  & \multicolumn{1}{l|}{  Yes }  & \multicolumn{1}{l|}{Imperceptible}             &   \multicolumn{1}{l|}{Imperceptible Nuis.}             & \multicolumn{1}{l|}{Imperceptible Nuis.}             & \multicolumn{1}{l|}{Managed Imperceptible}          \\ \cline{3-7}
\multicolumn{1}{|c|}{}                                   & \multicolumn{1}{l|}{}                                      & \multicolumn{1}{l|}{  No } & \multicolumn{1}{l|}{Imperceptible Ignored }     & \multicolumn{1}{l|}{Imper. Ignored Nuis.}     & \multicolumn{1}{l|}{Imper. Ignored Nuis.}     & \multicolumn{1}{l|}{Managed Imperceptible} \\ \hline
\multicolumn{1}{|c|}{\multirow{4}{*}{\rotatebox[origin=c]{90}{{No}}}} &
\multicolumn{1}{l|}{\multirow{2}{*}{\rotatebox[origin=c]{90} {Yes}}} 
  & \multicolumn{1}{l|}{  Yes}  & \multicolumn{1}{l|}{Faux}                     & \multicolumn{1}{l|}{Faux Nuis.}                     & \multicolumn{1}{l|}{Faux Nuis.}                     & \multicolumn{1}{l|}{Managed Faux}                  \\ \cline{3-7} 
\multicolumn{1}{|l|}{}                                   & \multicolumn{1}{l|}{}                                      & \multicolumn{1}{l|}{  No} & \multicolumn{1}{l|}{Faux  Ignored}            & \multicolumn{1}{l|}{Faux  Ignored Nuis.}            & \multicolumn{1}{l|}{Faux  Ignored  Nuis.}            & \multicolumn{1}{l|}{Managed Faux}          \\ \cline{2-7} 
\multicolumn{1}{|l|}{}                                   &
\multicolumn{1}{l|}{\multirow{2}{*}{\rotatebox[origin=c]{90}{No   }}}
  & \multicolumn{1}{l|}{Yes}  & \multicolumn{1}{l|}{Faux}                     & \multicolumn{1}{l|}{Faux  Nuis.}                     & \multicolumn{1}{l|}{Faux  Nuis.}                     & \multicolumn{1}{l|}{Managed Faux}                  \\ \cline{3-7} 
\multicolumn{1}{|l|}{}                                   & \multicolumn{1}{l|}{}                                      & \multicolumn{1}{l|}{No} & \multicolumn{1}{l|}{\parbox{1.in}{ \vspace{2pt} No novelty } } & \multicolumn{1}{l|}{No novelty  Nuis. } & \multicolumn{1}{l|}{{\parbox{1.in}{ \vspace{2pt} No novelty  Nuis. } } } & \multicolumn{1}{l|}{No Novelty}                    \\ \hline
\end{tabular}
\caption{Subtypes of novelty defined by interaction of primary novelty types and regret. 
Some combinations of states get multiple labels (\textit{e.g.}, Unanimous  Nuisance is both Unanimous (all types of novelty present) and Nuisance (inconsistent regret values)).
}
\label{tab:types}
\end{table*}
\renewcommand{\arraystretch}{1}
}

\comment{
\begin{figure}
 \centering
   \includegraphics[width=.9\columnwidth]{novelty_venn_diagram.pdf}
   \caption{Diagram highlighting the formally defined types of novelty and the relation to the primary world-level novelty, perceptual-level novelty, and agent-level novelty. 
   The black dashed line indicates the area of faux novelty.
  \label{fig:types}
 }
\end{figure}
}

\section{Types of Novelty} 
For the definitions, we introduce the primary types of novelty, with the subtypes defined by combining primary types and regret (see Table~\ref{tab:types} and \cite{Boult-eta-al-novelty20}).
Using our definitions of world, observation space, internal states, and the corresponding dissimilarity functions, we can formally define the following primary types of novelty.  
%


{\bf World novelty.} A world state $\check{w}\in{\cal W}$ is considered a world novelty for an agent $\alpha$ at time $t$ if $\min_{w \in E_{w,t}} {\cal D}_{w,{\cal T}} (w,\check{w}; E_{t}) > \delta_w$. That is, any world state $\check{w}$ sufficiently dissimilar from every world state in the experience tensor is a world-level novelty. In general, only an oracle with access to $E_{w,t}$ and ${\cal D}_{w,{\cal T}}$ can determine that a world state is truly novel. If the world representation is viewed as including distributional information (\textit{e.g.}, probabilities of various items occurring) then a change in distributional parameters can be a world-level novelty even if no new ``objects'' occur in the world. Thus world-level novelty can produce problems of domain adaption, not just domain transfer.

{\bf Observation novelty.} A world state $\check{w}\in{\cal W}$ is considered an observation novelty for an agent $\alpha$ at time $t$ iff $\min_{w \in E_{w,t}} {\cal D}_{o,{\cal T}} (w,\check{w};E_t) > \delta_o$.
That is, an observation novelty is the observation-space state obtained for any world state $\check{w}$ that, when projected through a perceptual operator, is sufficiently dissimilar from every observation-space state in the agent's experience tensor.
Note that in this definition, the observed world state, $\check{w}$, is subject to the current perceptual operator ${\cal P}$ at time $t$ and is compared to the observation-space states in the experience tensor, which may have used a potentially different perceptual operator ${\cal P}$.
It is not surprising that the same world state may be novel at one point in time but not novel at another. However, it may be surprising that if the perceptual operator changes over time, then something can be perceptually novel at time $t$ even if it was not perceptually novel at time $t-1$. For example, consider a transmission glitch creating errors in a static scene that has been viewed previously.
It is important to note that observation novelty is defined considering all experience, which permits observation novelty that includes distributional shifts or reasoning between states to detect novelty that impact dynamics.  
If the agent had access to the true dissimilarity ${\cal D}_{o,{\cal T}}$, it could use that to define its state recognition function $f$.  However, in practice, an agent will not have access to ${\cal D}_{o,{\cal T}}$, since it is trying to learn such a function from the data or was programmed with static rules to approximate it. Furthermore, agents rarely store all inputs.

{\bf Agent novelty.} An observation-space state $x={\cal P}(\check{w}\in{\cal W})$ is considered an agent novelty for an agent $\alpha$ at time $t$ {\em iff} $f_t(x) = {\cal N}$. That is, $x$ is an agent novelty {\em iff} the agent at time $t$ cannot map $x$ to any of its internal states or maps to a special state for when it detects novel inputs.  We note that this definition does not consider something novel if the state recognition functions $f_t$ associate $x$ with an incorrect state.


These novelty types are not mutually exclusive, and their combinations define the following notable novelty sub-types:
\begin{itemize}
    \item {\bf Unanimous novelty} is any world novelty $w$ for which the perceptual operator produces an observation-space state that is both an observation novelty and an agent novelty. Unanimous novelty is correctly detected by the agent. 
    \item {\bf Imperceptible novelty} is any world novelty $w$ for which the perceptual operator produces an observation space state $x$ that is not an observation novelty. Accordingly, the agent cannot directly react to such novelties. 
    \item {\bf Faux novelty} is a world state $w$ that is not a world novelty but its corresponding observation state $x$ is an observation novelty or an agent novelty. 
    \item {\bf Ignored novelty} is any world state $w$ such that its corresponding observation state $x$ is not an agent novelty while either $w$ is a world novelty or $x$ is an observation novelty. Ignored novelty does not have to result in poor performance (\textit{e.g.}, a non-adaptive agent may ignore all novelties while still performing well in the presence of them). 
\end{itemize}

Combining these novelty types and sub-types with the regret functions (${\cal R}_{f,{\cal T}}$, ${\cal R}_{o,{\cal T}}$, and ${\cal R}_{w,{\cal T}}$) allows us to formally define additional useful novelty sub-types including:
\begin{itemize}
    \item {\bf Managed novelty} is a world novelty $w$ such that
    its implication on regret (performance) is minimal, \textit{i.e.}, ${\cal R}_{f,{\cal T}}(w)<\epsilon$. 
    \item {\bf Nuisance novelty} is a novelty for which the world regret and the observation regret significantly disagree. 
\end{itemize}
These are important for evaluations defining  novelty ground-truth and associated world-regret, these sub-types need to be avoided or at least accounted for in evaluation metrics.

\subsection{Novelty Types in CartPole}

In the previous section, we have defined and measured dissimilarity and regret in the observation space. It is straightforward to develop the corresponding definitions in the world space, \textit{i.e.}, world dissimilarity shall include the unobserved parts of the world, in addition to observed parts, while world regret shall depend on the unobserved parts of the world.

These notions of dissimilarity and regret can be used to categorize novelty types.
The novelties discussed above in the CartPole section are both world and observation novelties, since both world and perceptual dissimilarities are larger than zero, ${\cal D}_{o,{\cal T}} (\check{x}_t, x_t) >0$ and ${\cal D}_{w,{\cal T}} (\check{w}_t, w_t) >0$.
%
%

If the CartPole environment had an additional unobserved cart that does not influence the main cart and pole, then any novelty in that unobserved cart would be an imperceptible world novelty since it would not influence transitions between the observed states, ${\cal D}_{o,{\cal T}} (\check{x}_t, x_t) =0$ and ${\cal D}_{w,{\cal T}} (\check{w}_t, w_t) >0$. If task performance (world regret) in either of these last two  also required detection of novelty, then they would be an imperceptible nuisance novelty.